\documentclass[a4paper,11pt]{article}
\usepackage{graphicx}
\usepackage{hyperref}

\usepackage{tikz}
\usepackage{comment}
\usepackage{amsmath,amssymb} 
\usepackage{color}

\usepackage[margin=1in]{geometry}

\usepackage{bbm}
\usepackage{booktabs}
\usepackage{algorithmic}
\usepackage{algorithm}
\usepackage{multirow}

\usepackage{amsmath,amsfonts,bm}
\usepackage{amssymb}
\makeatletter
\def\blfootnote{\xdef\@thefnmark{}\@footnotetext}
\makeatother

\newcommand{\mc}[1]{\mathcal{#1}}
\newcommand{\bb}[1]{\mathbb{#1}}

\def\eqref#1{equation~\ref{#1}}

\def\1{\bm{1}}
\newcommand{\train}{\mathcal{D}}

\def\rvx{{\mathbf{x}}}
\def\rvy{{\mathbf{y}}}

\def\vzero{{\bm{0}}}

\def\vtheta{{\bm{\theta}}}

\def\vx{{\bm{x}}}
\def\vy{{\bm{y}}}

\DeclareMathAlphabet{\mathsfit}{\encodingdefault}{\sfdefault}{m}{sl}
\SetMathAlphabet{\mathsfit}{bold}{\encodingdefault}{\sfdefault}{bx}{n}

\def\gB{{\mathcal{B}}}

\def\gD{{\mathcal{D}}}

\def\gG{{\mathcal{G}}}
\def\gH{{\mathcal{H}}}

\def\gX{{\mathcal{X}}}
\def\gY{{\mathcal{Y}}}

\newcommand{\KL}{D_{\mathrm{KL}}}

\title{\textbf{Robust and On-the-fly Dataset Denoising \\ for Image Classification}} 

\author{Jiaming Song 
\\ \texttt{tsong@cs.stanford.edu}
\\ Stanford University
\and
Lunjia Hu
\\ \texttt{lunjia@stanford.edu}
\\ Stanford University
\and 
Michael Auli 
\\ \texttt{michaelauli@fb.com}
\\ Facebook AI Research
\and
Yann Dauphin
\\ \texttt{yann@dauphin.io}
\\ Google Brain
\and 
Tengyu Ma 
\\ \texttt{tengyuma@stanford.edu}
\\ Stanford University
}
\begin{document}
\maketitle

\begin{abstract}
Memorization in over-parameterized neural networks could severely hurt generalization in the presence of mislabeled examples. However, mislabeled examples are hard to avoid in extremely large datasets collected with weak supervision.
We address this problem by reasoning counterfactually about the loss distribution of examples with uniform random labels had they were trained with the real examples, and use this information to remove noisy examples from the training set.
First, we observe that examples with uniform random labels have higher losses when trained with stochastic gradient descent under large learning rates. 
Then, we propose to model the loss distribution of the counterfactual examples using only the network parameters, which is able to model such examples with remarkable success. Finally, we propose to remove examples whose loss exceeds a certain quantile of the modeled loss distribution.
This leads to {On-the-fly Data Denoising} (\textsc{ODD}), a simple yet effective algorithm that is robust to mislabeled examples, while introducing almost zero computational overhead compared to standard training.
\textsc{ODD} is able to achieve state-of-the-art results on a wide range of datasets including real-world ones such as WebVision and Clothing1M.
\end{abstract}

\setlength{\tabcolsep}{1.4pt}

\section{Introduction}

\newcommand{\tnote}[1]{{\color{blue}Tengyu notes: #1}}

Over-parametrized deep neural networks have remarkable generalization properties while achieving near-zero training error~\cite{zhang2016understanding}. However, the ability to fit the entire training set is highly undesirable, as a small portion of mislabeled examples in the dataset could severely hurt generalization~\cite{zhang2016understanding,arpit2017a}.\blfootnote{Work done at Facebook AI research.}

Meanwhile, an exponential growth in training data size is required to linearly improve generalization in vision~\cite{sun2017revisiting}; this progress could be hindered if there are mislabeled examples within the dataset.

Mislabeled examples are to be expected in large datasets that contain millions of examples. Web-based supervision produces noisy labels~\cite{li2017webvision,mahajan2018exploring} whereas human labeled datasets sacrifice accuracy for scalability~\cite{krishna2016embracing}. 
Therefore, algorithms that are robust to various levels of mislabeled examples are warranted in order to further improve generalization for very large labeled datasets.

In this paper, we are motivated by the observation that crowd-sourcing or web-supervision could have multiple disagreeing sources; in such cases, noisy labels could exhibit higher conditional entropy than the ground truth labels. 
Since the information about the noisy labels (such as the amount of noise) is often scarce, we pursue \textit{a general approach} by counterfactual reasoning of the behavior of noisy examples with high conditional entropy.
Specifically, we reason about the \textit{counterfactual} case of how examples with uniform random noise would behave \textit{had they appeared in the training dataset}, without actually training on such labels. If a real example has higher loss than what most counterfactual examples with uniform random noise would have, then there is reason to believe that this example is likely to contain a noisy label; removing this example would then improve performance on a clean test set.

To reason about the \textit{counterfactual loss distribution} of examples with uniform random noise, we first show that training residual networks with \textit{large learning rates} will create a significant gap between the losses of clean examples and noisy examples. The distribution of training loss over clean examples decrease yet that of the uniformly noisy examples does not change, regardless of the proportion of noisy examples in the dataset.
Based on this observation, we propose a distribution that simulates the loss distribution of uniform noisy examples based only on the network parameters. Reasonable thresholds can be derived from percentiles of this distribution, which we can then utilize to denoise the dataset. This is critical in real-world applications, because prior knowledge about the distribution of label noise is often scarce; even if we have such information (such as transition matrices of label noise), algorithms that specifically utilize this information are not scalable when there are thousands of labels.

\begin{figure*}[t]
    \centering
    \includegraphics[width=\textwidth]{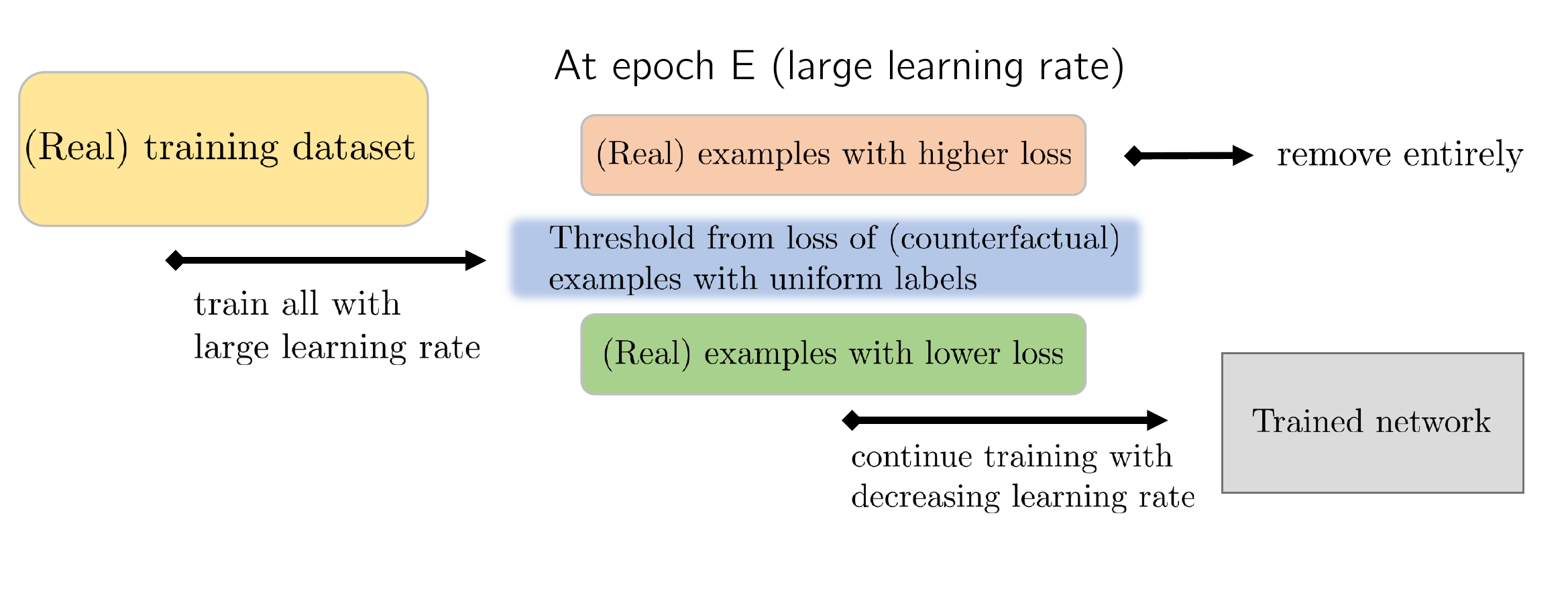}
    \caption{Pipeline of our method. We utilize the implicit regularization effect of SGD to (counterfactually) reason the loss distribution of examples with uniform label noise. We remove examples that have loss higher than the threshold and train on the remaining examples. There is no assumption that the dataset has to contain uniformly random labels (thus such labels are ``counterfactual''); we empirically validate our method on real-world noisy datasets.}
    \label{fig:pipeline}
\end{figure*}

\begin{figure*}[t]
    \centering
    \includegraphics[width=\textwidth]{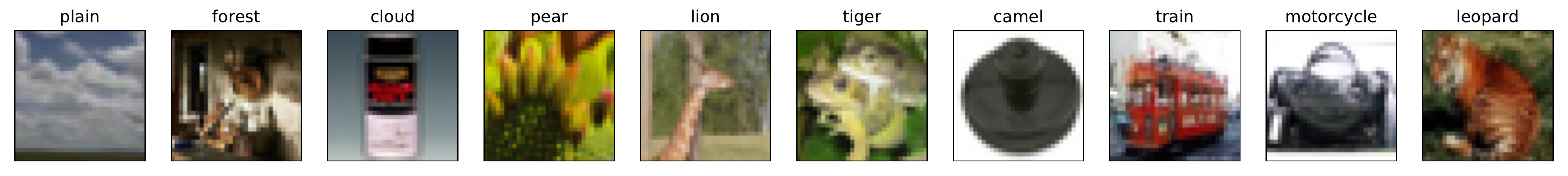}
    \caption{Mislabeled examples in the CIFAR-100 training set detected by \textsc{ODD}.}
    \label{fig:cifar-demo}
\end{figure*}

We proceed to propose \textit{On-the-fly Data Denoising} (\textsc{ODD}, see Figure~\ref{fig:pipeline}), a simple and robust method for training with noisy examples based on the implicit regularization effect of stochastic gradient descent.
First, we train residual networks with large learning rate schedules and use the resulting losses to separate clean examples from mislabeled ones. 
This is done by identifying examples whose losses exceed a certain threshold. 
Finally, we remove these examples from the dataset and continue training until convergence. \textsc{ODD} is a general approach that can be used to train clean dataset as well as noisy datasets with almost no modifications.

Empirically, \textsc{ODD} performs favorably against previous methods in datasets containing \textit{real-world noisy examples}, such as WebVision~\cite{li2017webvision} and Clothing1M~\cite{xiao2015learning}. 
\textsc{ODD} also achieves equal or better accuracy than the state-of-the-art \textit{on clean datasets}, such as CIFAR and ImageNet. 
We further conduct ablation studies to demonstrate that \textsc{ODD} is robust to different hyperparameters and artificial noise levels. Qualitatively, we demonstrate the effectiveness of \textsc{ODD} by detecting mislabeled examples in the ``clean'' CIFAR-100 dataset without any supervision other than the training labels (Figure~\ref{fig:cifar-demo}). These results suggest that we can use \textsc{ODD} in both clean and noisy datasets with minimum computational overhead to the training algorithm. 


\section{Problem setup}

The goal of supervised learning is to find a function $f \in \mc{F}$ that describes the probability of a random label vector $Y \in \mc{Y}$ given a random input vector $X \in \mc{X}$, which has underlying joint distribution $P(X, Y)$. Given a loss function $\ell(\rvy, \hat{\rvy})$, one could minimize the average of $\ell$ over $P$:
\begin{align}
\mc{R}(f) = \int \ell(\rvy, f(\rvx)) \ \mathrm{d} P(\rvx, \rvy), \label{eq:erm}
\end{align}
which is the basis of empirical risk minimization (\textsc{ERM})
The joint distribution $P(X, Y)$ is usually unknown, but we could gain access to its samples via a potentially noisy labeling process, such as crowdsourcing~\cite{krishna2016embracing} or web queries~\cite{li2017webvision}. 

We denote the training dataset with $N$ examples as $\train = (\rvx_i, \rvy_i)_{i \in [N]} = \mc{G} \cup \mc{B}$. $\mc{G}$ represents correctly labeled (clean) examples sampled from $P(X, Y)$. $\mc{B}$ represents mislabeled examples that are not sampled from $P(X, Y)$, but from another distribution $Q(X, Y)$; $\mc{G} \cap \mc{B} = \varnothing$, as a sample cannot be both correctly labeled and mislabeled. 

We aim to learn the function $f$ from $\train$ \emph{without knowledge about $\mc{B}$, $\mc{G}$ or their statistics} (e.g. $|\mc{B}|$). 
A typical approach is to pretend that $\mc{B} = \varnothing$ --- i.e., all examples are i.i.d. from $P(X, Y)$ --- and minimize the empirical risk:
$$
\hat{\mc{R}}(f) = \frac{1}{N} \sum_{i = 1}^{N} \ell(\rvy, f(\rvx)).
$$
If $\mc{B} = \varnothing$ is indeed true, then the empirical risk converges to the population risk: $\hat{\mc{R}}(f) \to \mc{R}(f)$ as $N \to \infty$. However, if $\mc{B} \neq \varnothing$, then $\hat{\mc{R}}(f)$ is no longer an unbiased estimator of $\mc{R}(f)$. 
Moreover, when $\mc{F}$ contains large neural nets with the number of parameters exceeding $N$, the empirical risk minimizer could fit the entire training dataset, including the mislabeled examples~\cite{zhang2016understanding}. Overfitting to wrong labels empirically causes poor generalization.  
For example, training CIFAR-10 with 20\% of uniformly mislabeled examples and a residual network gives a test error of 11.5\%, which is significantly higher than the 4.25\% error obtained with training on the clean examples\footnote{See Table.~\ref{tab:cifar-val}, Section~\ref{sec:exp-cifar} for the exact experiment setup.}.

\subsection{Entropy-based Assumption over Noisy Labels}
Therefore, if we were able to identify the clean examples belonging to $\mc{G}$, we could vastly improve the generalization on $P(X, Y)$; this requires us to provide valid prior assumptions that could distinguish clean examples from mislabeled ones. We note that these assumptions have to be general enough so as to \textit{not depend on additional assumptions specific to each dataset}. For example, knowledge about noise transition matrices is not allowed.

We assume that for any example $\vx \in \gX$, the entropy of the clean label distribution is smaller than that of the noisy label distribution:
\begin{align}
    \gH(P(Y | X = \vx)) < \gH(Q(Y | X = \vx)) \quad \forall \vx \in \gX
\end{align}
where the randomness of labeling $Q(Y | X)$ could arise from noisy labelings, such as Mechanical Turk~\cite{krishna2016embracing}.
Let $\ell$ be the cross entropy loss, then the ERM objective is essentially trying to minimize the KL divergence between the empirical conditional distribution (denoted as $\hat{P}(\vy | \vx)$) and the conditional distribution parametrized by our model (denoted as $p_\vtheta(\vy | \vx)$):
\begin{align}
    \bb{E}_{\hat{P}(\vy | \vx)} [-\log p_\vtheta(\vy | \vx)] = \gH(\hat{P}(\vy |\vx)) + \KL(\hat{P}(\vy | \vx) \Vert p_\vtheta(\vy | \vx))
\end{align}
which is minimized as $\KL \to 0$; in this case, the cross entropy loss is higher if $\hat{P}$ has higher entropy, which suggests that the mislabeled examples are likely to have higher loss than correct ones.

\section{Denoising datasets on-the-fly with Counterfactual Thresholds}
In the following section, we study the behavior of samples with uniformly random label noise; this allows us to reason about their loss distribution \textit{counterfactually}, and develop suitable thresholds to remove noisy examples that appear in the training set. 

The conditional distribution $Q_U(Y|X)$ of uniformly random label noise is simply:
\begin{align}
    Q_U(Y|X) = \mathrm{Uniform}(\gY).
\end{align}
We note that $Q_U(Y|X)$ is the distribution that maximizes entropy; therefore, any real-world noise distribution $Q(Y|X)$ will have smaller entropy than $Q_U$. 

While it is unreasonable to assume that the label noise is uniformly random in practice, we do not make such assumption over our training set. Instead, we reason about the following counterfactual case: 
\begin{quote}
    \textit{Had the training set contained some examples with uniform random labels, can we characterize the loss distribution of these examples?}
\end{quote}

Then, we illustrate how such a counterfactual analysis allows simple and practical algorithms that work even under real-world noisy datasets.
\begin{itemize}
    \item First, we show that when training ResNets via SGD with large learning rates, the training loss of uniform noisy labels and clean labels can be clearly separated.
    \item Next, we propose an approach to model the (counterfactual) loss distribution 
    \textit{by only looking at the weights of the network}. We empirically show that this does not depend on the type or the amount of noisy labels in the dataset, making this approach generalize well to various counterfactual scenarios (such as different portions of uniform random labels in the dataset).
    \item Finally, we can simply remove all examples that perform worse than a certain percentile of the counterfactual distribution. Since higher entropy examples tend to have higher loss than lower entropy ones, the samples we remove are more likely to be more noisy. In Fig.~\ref{fig:cifar-demo}, we empirically demonstrate that the proposed threshold identifies mislabeled samples in CIFAR-100 even without any additional supervision, validating our assumption.
\end{itemize}

\begin{figure*}[t]
\begin{center}
\includegraphics[width=0.9\textwidth]{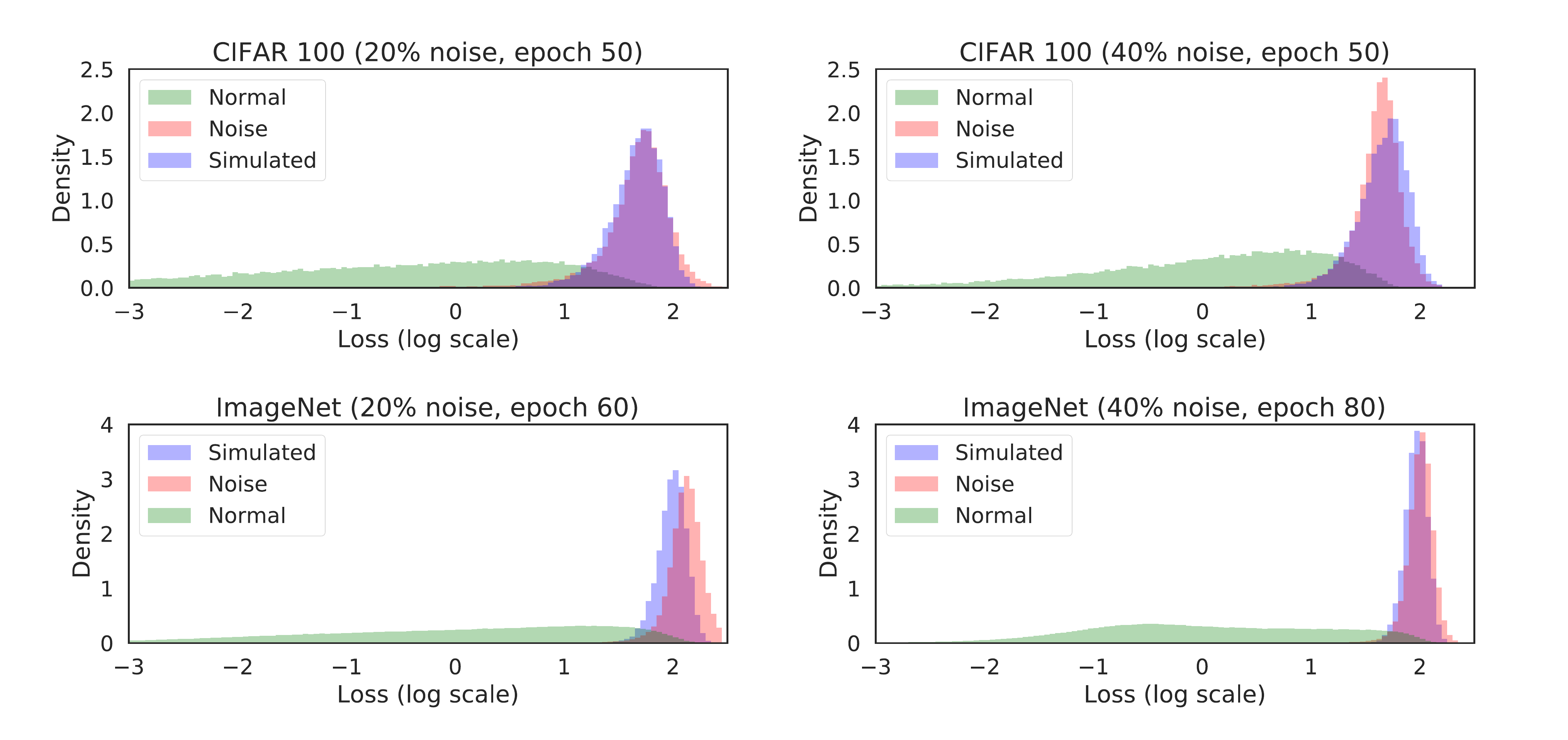}
\caption{Histogram of the distributions of losses, where ``normal'', ``noise'', and ``simulated'' denote (real) examples with clean labels, (real) examples with uniform random labels and the counterfactual model $q_n(\ell)$ respectively. 
$q_n(\ell)$ matches the loss distribution of noisy examples, which have higher loss than clean ones; $q_n(\ell)$ depends only on the network parameters.}
\label{fig:histogram}
\end{center}
\end{figure*}



\subsection{Separating mislabeled examples via SGD}

First, we find that training the model with stochastic gradient descent (SGD) with large learning rates (e.g. $0.1$) will result in significant discrepancy between the loss statistics of the clean examples and mislabeled examples. We consider training deep residual networks on CIFAR-100 and ImageNet with different percentages of uniform label noise ($20\%$ and $40\%$), but with large learning rates (close to $0.1$), and at specific epochs, we plot the histogram of the loss for each example. 

As demonstrated in Fig.~\ref{fig:histogram}, the loss distributions of clean examples and mislabeled ones have notable statistical distance. Moreover, it seems that the loss distribution of the uniform labeled examples are relatively stable, and \textit{does not depend on the amount of uniform random noise in the training set}.
This is consistent with the obeservations in~\cite{zhang2016understanding}, as the network starts to fit mislabeled examples when learning rate decrease further; decreasing learning rate is crucial for achieving better generalization on clean datasets.

The working of the implicit regularization of stochastic gradient descent is by and large an open question that attracts much recent attentions~\cite{neyshabur2017implicit,li2017algorithmic,du2018algorithmic,mandt2017stochastic}. Empirically, it has been observed that large learning rates are beneficial for generalization~\cite{kleinberg2018an}. Chaudhari and Soatto~\cite{chaudhari2018stochastic} have argued that SGD iterates converge to limit cycles with entropic regularization proportional to the learning rate and inversely proportional to batch size. Training with large learning rates under fixed batch sizes could then encourage solutions that are more robust to large random perturbations in the parameter space and less likely to overfit to mislabeled examples.

Given these empirical and theoretical evidences on large learning rate helps generalization, we propose to classify correct and mislabeled examples through the loss statistics, and achieve better generalization by removing the examples that are potentially mislabeled.



\subsection{Thresholds that classify mislabeled examples}
The above observation suggests that it is possible to distinguish clean and noisy examples \textit{via a threshold over the loss value}. In principle, we can claim an example is noisy if its loss value exceeds a certain threshold; by removing the noisy labels from the training set, we could then improve generalization performance on clean validation sets.

However, to improve generalization in practice, one critical problem is to select a reasonable threshold for classification. High thresholds could include too many examples from $\mc{B}$ (the mislabeled set), whereas low thresholds could prune too many examples from $\mc{G}$ (the clean set); reasonable thresholds should also adapt to different ratios of mislabeled examples, which could be unknown to practitioners. 


If we are able to characterize the loss of $Q_U(Y|X)$ (the highest entropy distribution), we can select a reasonable threshold from this loss as any example having higher loss is likely to have high entropy labels (and is possibly mislabeled).
From Fig.~\ref{fig:histogram}, the loss distribution for $\mc{B}$ is relatively stable with different ratios of $|\mc{B}|/|\mc{D}|$; examples in $\mc{B}$ are making little progress when learning rate is large. This suggests a threshold selecting criteria that is \textit{independent of the amount of mislabeled examples in the dataset}.

We propose to characterize the loss distribution of (counterfactual) uniform label noise via the following procedure:
\begin{gather}
l =  -\tilde{\rvy}_k + \log \left(\sum_{i \in [N]} \exp(\tilde{\rvy}_i)\right) \label{eq:loss-dist} \\
\tilde{\rvy} = \mathrm{fc}(\mathrm{relu}(\tilde{\rvx})), \tilde{\rvx} \sim \mathcal{N}(0, I), k \sim \mathrm{Uniform}\{0, \ldots, K\} \nonumber
\end{gather}
We denote this counterfactual distribution model as $q_n(l)$.

$q_n(l)$ tries to simulate the behavior of the model (and the loss distribution) with several components.
\begin{itemize}
    \item $k$ represents a random label from $K$ classes. This simulates the case where $Q(Y|X)$ has the highest entropy, i.e. uniformly random.
    \item $\mathrm{fc}(\cdot)$ is the final (fully connected) layer of the network and $\mathrm{relu}(\tilde{\rvx}) = \max(\tilde{\rvx}, \vzero)$ is the Rectified Linear Unit. This simulates the behavior at the last layer of the network outputs $\tilde{\vy}$.
    \item $\tilde{\rvx} \sim \mathcal{N}(0, I)$ suggests that the inputs to the last layer has an identity covariance; the scale of the covariance could result from well-conditioned objectives defined via deep residual networks~\cite{he2015deep}, batch normalization~\cite{ioffe2015batch} and careful initialization~\cite{he2015delving}.
\end{itemize}
We qualitatively demonstrate the validity of our characterization on CIFAR-100 and ImageNet datasets in Fig.~\ref{fig:histogram}, where we plot the histogram of the $q_n(l)$ distribution for CIFAR-100 and ImageNet, and compare then with the empirical distribution of the loss of uniform noisy labeled examples.
The similarities between the noisy loss distribution and simulated loss distribution $q_n(l)$ demonstrate that an accurate characterization of the loss distribution can be made \textit{without prior knowledge of the mislabeled examples}. 

To effectively trade-off between precision (correctly identifying noisy examples) and recall (identifying more noisy examples), we define a threshold via the $p$-th percentile of $q_n(l)$ using the samples generated by Equation~\ref{eq:loss-dist}; it relates to approximately how much examples in $\mc{B}$ we would retain if $Q(Y|X)$ is uniform. In Section~\ref{sec:exp-ablation}, we show that this method is able to identify different percentages of uniform label noise with high precision.


\subsection{A Practical Algorithm for Robust Training}
We can utilize this to remove examples that might harm generalization, leading to \textit{On-the-fly Data Denoising} (\textsc{ODD}), a simple algorithm robust to mislabeled examples.

\begin{algorithm}
  \caption{On-the-fly Data Denoising}
  \label{alg:odd}
\begin{algorithmic}
  \STATE {\bfseries Input:} dataset $\gD$ of size $N$, model $f_\theta$, percentile $p$, epoch $E$, learning rate schedule $\eta(t)$.
  \FOR{$e = 1 \dots E$}
  \STATE Train on $\gD$ with learning rate $\eta(e)$.
  \ENDFOR
  \STATE $T_p$ = $p$-th percentile of $q_n(\ell)$ in Eq.~(\ref{eq:loss-dist}) 
  \STATE $\gG = \{(\vx, \vy)| \ell(\vy, f_\theta(\vx)) < T_p\}$, $\gB = \gD \setminus \gG$.
  \FOR{$e = E+1 \dots $}
  \STATE Train on $\gG$ with learning rate $\eta(e)$.
  \ENDFOR
\end{algorithmic}
\end{algorithm}


\subsubsection{Hyperparameter selection} \textsc{ODD} introduces two hyperparameters: $E$ determines the amount of training that separates clean examples from noisy ones; $p$ determines $T_p$ that specifies the trade-off between less noisy examples and more clean examples. We do not explicitly estimate the portion of noise in the dataset, nor do we assume any specific noise model. 
Moreover, \textsc{ODD} is compatible with existing practices for learning rate schedules, such as stepwise~\cite{he2015deep} or cosine~\cite{loshchilov2016sgdr}.

\begin{figure}
    \centering
    \includegraphics[width=\textwidth]{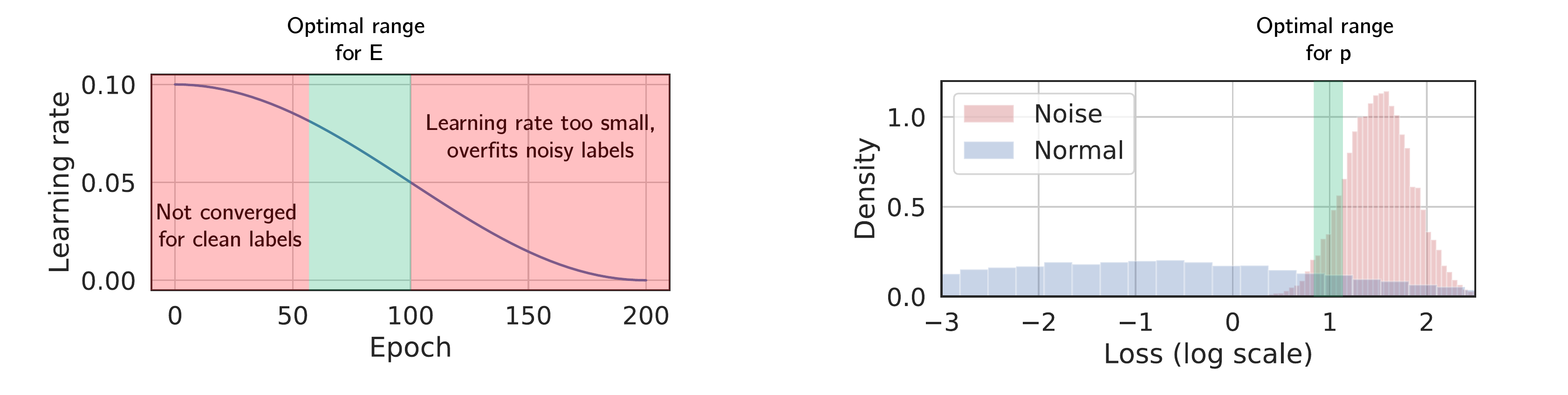}
    \caption{Hyperparameter selection. (Left) Cosine learning rate schedule across epochs; we wish to select $E$ before learning rate becomes small, and after training over clean labels have converged. (Right) Histogram of the losses; we wish to select $p$ that does not remove too many clean data, but also removes as many (conterfactually) noisy data as possible.}
    \label{fig:hyper}
\end{figure}

In Fig.~\ref{fig:hyper} we demonstrate and discuss how to choose the hyperparameters $E$ and $p$. For $E$, we wish to perform ODD operation at a point not too early (to allow enough time for training on clean labels to converge) and not too late (to prevent overfitting noisy labels with small learning rates). For $p$, we wish to trade-off between keeping as much clean data as possible and removing counterfactually noisy data; selecting $p \in [1, 30]$ typically works for our case.
\section{Experiments}
\label{sec:experiments}
We evaluate our method extensively on several clean and noisy datasets including CIFAR-10, CIFAR-100,  ImageNet~\cite{russakovsky2015imagenet}, WebVision~\cite{li2017webvision} and Clothing1M~\cite{xiao2015learning}. CIFAR-10, CIFAR-100 and ImageNet are clean whereas WebVision and Clothing1M are obtained via web supervision and have more noisy labels. Our experiments consider datasets that are clean, have artificial noise (in CIFAR-10, CIFAR-100 and ImageNet), or have inherent noise from web-supervision (as in the case of WebVision and Clothing1M). 




\begin{table*}[htbp]
\begin{center}
\caption{Validation accuracy (in percentage) with uniform label noise.}
\label{tab:cifar-val}
\begin{tabular}{c|ccc|ccc}
\toprule
& \multicolumn{3}{c|}{CIFAR-10} & \multicolumn{3}{c}{CIFAR-100} \\
  $\%$ mislabeled                                           & 0                              & 20                              & 40                & 0 & 20 & 40             \\\midrule
\textsc{ERM}                                           & $96.3 \pm 0.1$ & $88.5 \pm 0.1$ & $84.4 \pm 0.5$ &  $81.6 \pm 0.2$ & $69.6 \pm 0.1$ & $55.7 \pm 0.5$ \\
\textit{mixup}                                                &           ${97.0} \pm 0.1$     & $93.9 \pm 0.3$ & $91.7 \pm 0.1$  & $81.4 \pm 0.3$ & $71.2 \pm 0.3$ & $59.4 \pm 0.4$  \\
\textsc{GCE}     &  -  & $89.9 \pm 0.2$  & $87.1 \pm 0.2$  & - & $66.8 \pm 0.4$ & $62.7 \pm 0.2$ \\
\textsc{Luo} & $96.2 \pm 0.1 $ & $\textbf{96.2} \pm 0.2$ & $94.9 \pm 0.2$ & $81.4 \pm 0.2$ & $80.6 \pm 0.5$ & $74.2 \pm 0.5$ \\
\midrule
\textsc{Ren}$^\star$   &    -    &    -     & $86.9 \pm 0.2$ & - & - & $61.4 \pm 2.0$  \\
\textsc{MentorNet}$^\star$ &   -      & 92.0    & 89.0  & - & 73.0 & 68.0 \\\midrule
\textsc{ODD}                              & $96.2 \pm 0.1$ & ${94.7} \pm 0.1$  & ${92.8} \pm 0.2$ & ${81.8} \pm 0.1$ & ${77.2} \pm 0.1$  & ${72.4} \pm 0.4$    \\
\textsc{ODD} + \textit{mixup}                              & $\textbf{97.2} \pm 0.1$ & $\textbf{95.6} \pm 0.1$  & $\textbf{95.5} \pm 0.2$ & $\textbf{82.5} \pm 0.1$ & $\textbf{79.1} \pm 0.1$  & $\textbf{76.5} \pm 0.4$    \\
\bottomrule                          
\end{tabular}
\end{center}
\end{table*}

\subsection{CIFAR-10 and CIFAR-100}
\label{sec:exp-cifar}
We first evaluate our method on the CIFAR-10 and CIFAR-100 datasets, which contain 50,000 training images and 10,000 validation images of size $32 \times 32$ with 10 and 100 labels respectively. 
In our experiments, we train the wide residual network architecture (WRN-28-10) in~\cite{zagoruyko2016wide} for 200 epochs with a minibatch size of 128, momentum $0.9$ and weight decay $5 \times 10^{-4}$. We set $E = 75$ (total number of epochs is 200) and $p = 10$ in our experiments.

\subsubsection{Input-Agnostic Label Noise}

We first consider label noise that are agnostic to inputs. 
Following~\cite{zhang2016understanding}, We randomly replace a $0\% / 20\% / 40\%$) of the training labels to uniformly random ones, and evaluate generalization error on the clean validation set.
We compare with the following baselines: 
Empirical Risk Minimization (\textsc{ERM}, Eq.~\ref{eq:erm},~\cite{goyal2017accurate}) which assumes all examples are clean; \textsc{MentorNet}~\cite{jiang2017mentornet}, which pretrains an auxiliary model that predicts weights for each example based on its input features; \textsc{Ren}~\cite{ren2018learning}, which optimizes the weight of examples via meta-learning; 
\textit{mixup}~\cite{zhang2017mixup}, a data augmentation approach that trains neural networks on convex combinations of pairs of examples and their labels; Generalized Cross Entropy (\textsc{GCE},~\cite{zhang2018generalized}) that includes cross-entropy loss and mean absolute error~\cite{ghosh2017robust}; and \textsc{Luo}~\cite{luo2019simple}, which regularizes the Jacobian of the network. We also consider using \textit{mixup} training after we pruned noisy examples with \textsc{ODD}. 

\begin{figure*}[tbp]
    \centering
    \includegraphics[width=0.9\textwidth]{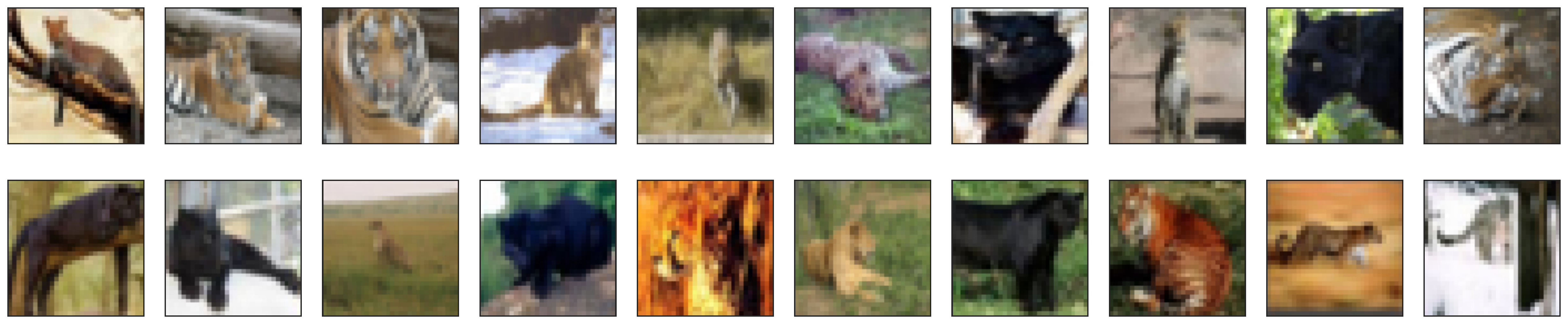}
    \caption{Examples with label ``leopard'' that are classified as mislabeled.}
    \label{fig:leopard}
\end{figure*}

\begin{figure*}[tbp]
    \centering
    \includegraphics[width=0.9\textwidth]{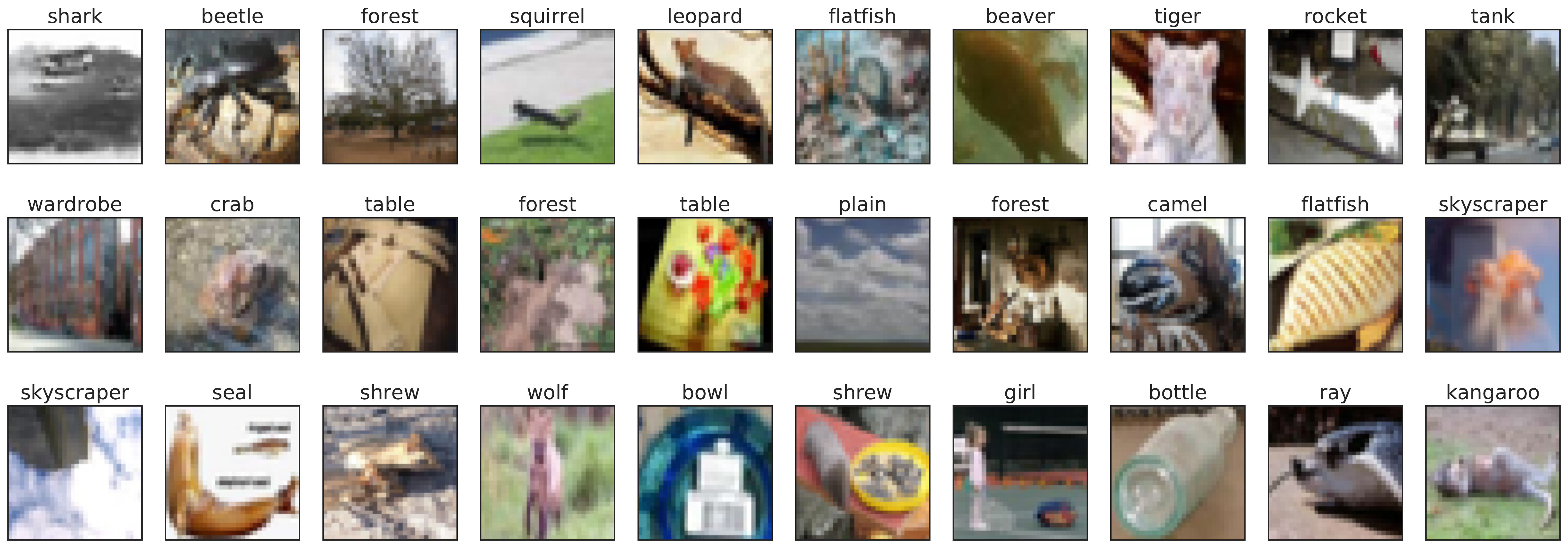}
    \caption{Random CIFAR-100 examples that are classified as mislabeled.}
    \label{fig:noisy-images-cifar100-t}
\end{figure*}

We report the top-1 validation error in Table~\ref{tab:cifar-val}, where $\star$ denotes methods trained with knowledge of 1000 additional clean labels. Notably, \textsc{ODD} + \textit{mixup} significantly outperforms all other algorithms (except for \textsc{LUO} with 20\% noise on CIFAR-10). On the one hand, this suggests that \textsc{ODD} is able to distinguish the mislabeled examples and improve generalization; on the other hand, it would seem that removing certain examples even in the ``clean'' dataset does not seem to hinder generalization, suggesting that our thresholds works in practice.



\subsubsection{Mislabeled examples in CIFAR-100} 
We display the examples in CIFAR-100 training set for which our \textsc{ODD} methods identify as noise across 3 random seeds. One of the most common label such examples have is ``leopard''; in fact, 21 of 50 ``leopard'' examples in the training set are perceived as hard, and we show some of them in Fig.~\ref{fig:leopard}. It turns out that a lot of the ``leopard'' examples contains images that clearly contains tigers and black panthers (CIFAR-100 has a label corresponding to ``tiger''). We also demonstrate random examples from the CIFAR-100 that are identified as noise in Fig.~\ref{fig:noisy-images-cifar100-t}. 
The examples identified as noise often contains multiple objects, or are more ambiguous in terms of identity. 
We include more results in Appendix~\ref{sec:app:noisy-cifar}.

\setlength{\tabcolsep}{5pt}
\begin{table}
\centering
    \caption{Results on non-homogeneous labels.}
\label{tab:cifar-merge}
\begin{tabular}{c|c|cc}
\toprule
    Task & \% samples removed ($c$) & \textsc{ERM} & \textsc{ODD} \\\midrule
  \multirow{3}{*}{CIFAR-50} 
  & 30 & $78.5 \pm 0.1$ & $\textbf{79.0} \pm 0.1$ \\
  & 50 & $77.9 \pm 0.1$ & $\textbf{78.6} \pm 0.2$ \\
  & 70 & $77.5 \pm 0.1$ &  $\textbf{78.1} \pm 0.1$ \\\midrule
  \multirow{3}{*}{CIFAR-20}  
  & 30 & $86.4 \pm 0.2$ & $\textbf{86.6} \pm 0.1$ \\
  & 50 & $85.1 \pm 0.1$ & $\textbf{85.4} \pm 0.2$ \\
  & 70 & $84.4 \pm 0.3$ & $\textbf{84.7} \pm 0.2$ \\
  \bottomrule
\end{tabular}
\end{table}


\begin{table}
\begin{center}
\caption{Top-1 (top-5) accuracy on ImageNet.}
\label{tab:imagenet-uniform-noise}
\begin{tabular}{c|ccc}
\toprule
$\%$ mislabeled &  0 & 20 & 40 \\\midrule
\textsc{ERM} & \textbf{78.7} (94.3) & 72.6 (90.2) & 61.2 (84.4) \\
\textsc{Luo} & 76.7 (93.3) & 75.2 (92.3) & 73.2 (91.0) \\
\textsc{MentorNet} & -  & -  & 65.1 (85.9) \\\midrule
\textsc{ODD} ($p=10$) & \textbf{78.7} (94.0) & \textbf{77.5} (93.5) & \textbf{74.8} (92.1)
  \\\bottomrule
\end{tabular}
\end{center}
\end{table}

\subsubsection{Non-Homogeneous Labels}
We evaluate \textsc{ERM} and \textsc{ODD} on a setting without mislabeled examples, but the ratio of classes could vary. To prevent the model from utilizing the number of examples in a class, we combine multiple classes of CIFAR-100 into a single class, creating the CIFAR-20 and CIFAR-50 tasks. In CIFAR-50, we combine an even class with an odd class while we remove $c\%$ of the examples in the odd class. In CIFAR-20, we combine 5 classes in CIFAR-100 that belong to the same super-class
while we remove $c\%$ of the examples in 4 out of 5 classes. This is performed for both training and validation datasets. Results for \textsc{ERM} and \textsc{ODD} with $p = 10$ and $E = 75$ are shown in Table~\ref{tab:cifar-merge}, where \textsc{ODD} is able to outperform \textsc{ERM} in these settings where the input examples are not uniformly distributed.



\subsection{ImageNet}
\label{sec:exp-imagenet}
We consider experiments on the ImageNet-2012 classification dataset~\cite{russakovsky2015imagenet}.  
Input-agnostic random noise of $0\%, 20\%, 40\%$ are considered. 
We only use the center $224 \times 224$ crop for validation. 
We train ResNet-152 models~\cite{he2015deep} 
for 
$90$ epochs and report top-1 and top-5 validation errors in Table~\ref{tab:imagenet-uniform-noise}. \textsc{ODD} significantly outperforms \textsc{ERM} and \textsc{Luo}~\cite{luo2019simple} in terms of both top-1 and top-5 errors with 20\% and 40\% label noise, while being comparable to \textsc{ERM} on the clean dataset.

\subsection{WebVision}
\label{sec:exp-webvision}

\begin{table}
\centering
\caption{Top-1 (top-5) accuracy on WebVision and ImageNet validation sets when trained on WebVision. 
}
\label{tab:webvision}
    \begin{tabular}{c | c c}
    \toprule
     Method    & WebVision & ImageNet \\\midrule
        
        LASS~\cite{arpit2017a} & 66.6 (85.6) & 59.0 (80.8) \\
        CleanNet~\cite{lee2018cleannet}  & 68.5 (86.5) & 60.2 (81.1) \\
        \textsc{ERM} & 69.7 (87.0) & 62.9 (83.6) \\
        \textsc{MentorNet}~\cite{jiang2017mentornet} & 70.8 (88.0) & 62.5 (83.0) \\
        CurriculumNet~\cite{guo2018curriculumnet} & 73.1 (89.2) & 64.7 (84.9) \\
        \textsc{Luo}~\cite{luo2019simple} & 73.4 (89.5) & 65.9 (85.7) \\
        \midrule
        \textsc{ODD} &  \textbf{74.6} (90.6) & \textbf{66.7} (86.3)  \\\bottomrule
    \end{tabular}
\end{table}

We further verify the effectiveness of our method on a real-world noisy dataset. The WebVision-2017 dataset~\cite{li2017webvision} contains 2.4 million of real-world noisy labels, that are crawled from Google and Flickr using the 1,000 labels from the ImageNet-2012 dataset. 
We consider training Inception ResNet-v2~\cite{szegedy2016inception} for 50 epochs and use input images of size $299 \times  299$. 
We use both WebVision and ImageNet validation sets for 1-crop validation, following the settings in~\cite{jiang2017mentornet}. We do not use a pretrained model or additional labeled data from ImageNet.
In Table~\ref{tab:webvision}, we demonstrate superior results than other competitive methods tailored for learning with noisy labels. 

Our \textsc{ODD} method with $p = 30$ 
removes $9.3\%$ of the total examples with Inception ResNet-v2~\cite{szegedy2016inception}. Table~\ref{tab:webvision} suggests that our method is able to outperform the baseline methods when the training dataset is noisy, even as we remove a notable portion of examples. 
In comparison, we removed around $1.1\%$ of examples in ImageNet;  
this suggest that WebVision labels are indeed much noisier than the ImageNet labels since there are more examples removed by the (counterfactual) threshold. 


\subsection{Clothing1M}
\label{sec:clothing1m}

\begin{table}
\centering
 \caption{Validation accuracy on Clothing1M.}
    \begin{tabular}{c|c|c}
    \toprule
        Method & Setting & Accuracy \\\midrule
        \textsc{ERM} & noisy & 68.9 \\
        GCE & noisy & 69.1 \\
        Loss Correction~\cite{patrini2017making} & noisy & 69.2 \\
        LCCN~\cite{yao2019safeguarded} & noisy & 71.6 \\
        Joint Opt.~\cite{tanaka2018joint} & noisy & 72.2 \\
        DMI~\cite{xu2019l_dmi} & noisy & 72.5 \\
        \textsc{ODD} & noisy & \textbf{73.5} \\\midrule
        \textsc{ERM} & clean & 75.2 \\
        Loss Correction & noisy + clean & \textbf{80.4} \\
        \textsc{ODD} & noisy + clean & \textbf{80.3} \\
        \bottomrule
    \end{tabular}
    \label{tab:clothing1m}
\end{table}

Clothing1M~\cite{xiao2015learning} contains 1 million examples with noisy labels and 50,000 examples with clean labels 
of 14 classes. Following procedures from previous work, we use the ResNet-50 architecture pre-trained on ImageNet, with a starting learning rate of 0.001 trained with 10 epochs. We consider three settings, where the dataset contains \textit{clean} labels only, \textit{noisy} labels only, or both types of labels. For \textsc{ODD}, we set $E = 1, p = 1$ for the noisy dataset ($E = 1$ because we fine-tune from ImageNet pre-trained model); we then fine-tune on the clean labels if they are available. 

Table~\ref{tab:clothing1m} suggests our method compares favorably against existing methods such as GCE, Joint Optimization~\cite{tanaka2018joint}, 
latent class-conditional noise
model (LCCN,~\cite{yao2019safeguarded}) and Determinant based Mutual
Information (DMI,~\cite{xu2019l_dmi}) on the \textit{noisy} dataset, and is comparable to Loss Correction (LC, ~\cite{patrini2017making}) on the \textit{noisy + clean dataset}. We note that LC estimates the label confusion matrix using examples with both clean and noisy labels; the complexity of LC scales quadratically in the number of classes, and it would not be feasible for ImageNet or WebVision.

\subsection{Ablation Studies}
\label{sec:exp-ablation}


We include additional ablation studies in Appendix~\ref{sec:exp-ablation}.

\begin{figure}
    \centering
    \includegraphics[width=0.9\textwidth]{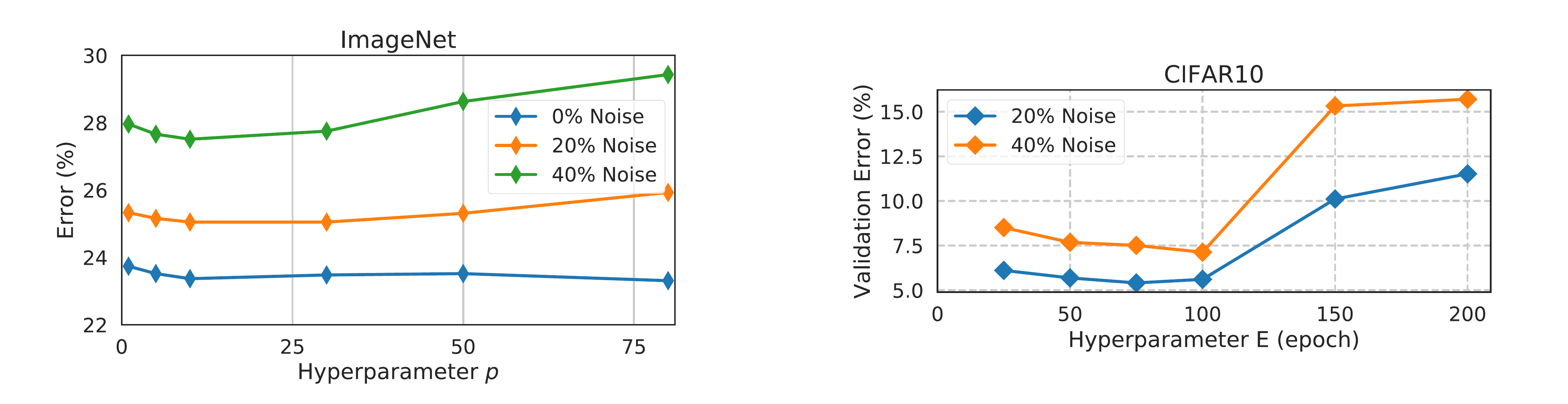}
    \caption{(Left) ablating $p$ on ImageNet. (Right) ablating $E$ on CIFAR10.}
    \label{fig:ablations}
\end{figure}

\subsubsection{Sensitivity to $p$} We first evaluate noisy ImageNet classification with varying $p$. 
A higher $p$ includes more clean examples at the cost of involving more noisy examples. 
From Fig.~\ref{fig:ablations} (left), \textsc{ODD} is not very sensitive to $p$, and empirically $p = 10$ represents the best trade-off. 

\subsubsection{Sensitivity to $E$} We evaluate the validation error of \textsc{ODD} on CIFAR with $20\%$ and $40\%$ input-agnoistic label noise where $E \in \{25, 50, 75, 100, 150, 200\}$ ($E = 200$ is equivalent to \textsc{ERM}). The results in Fig.~\ref{fig:ablations} (right) demonstrate that the effect of $E$ on final performance behaves according to our suggestion. 

\section{Related work}
\subsubsection{Generalization of SGD Training} The generalization of neural networks trained with SGD depend heavily on learning rate schedules~\cite{loshchilov2016sgdr}. It has been proposed that wide local minima 
could result in better generalization~\cite{hochreiter1995simplifying,chaudhari2016entropy,keskar2016on}. Several factors could contribute to wider local optima and better generalization, such as smaller minibatch sizes~\cite{keskar2016on}, reasonable learning rates~\cite{kleinberg2018an}, longer training time~\cite{hoffer2017train}, or distance from the initialization point~\cite{hoffer2017train}. 
In the presence of mislabeled examples, changes in optimization landscape~\cite{arpit2017a} could result in bad local minima~\cite{zhang2016understanding}, although it is argued that larger batch sizes could mitigate this effect~\cite{rolnick2017deep}.

\subsubsection{Training with Mislabeled Examples} One paradigm involves estimating the noise distribution~\cite{liu2014classification} or confusion matrix~\cite{sukhbaatar2014training}. 
Another line of methods propose to identify and clean the noisy examples~\cite{cretu2008casting} through predictions of auxillary networks~\cite{veit2017learning,patrini2017making} or via binary predictions~\cite{northcutt2017learning}; the noisy labels are either pruned~\cite{brodley1996identifying} or replaced with model predictions~\cite{reed2014training}. 
Our method is comparable to these approaches, but the key difference is that we leverage the implicit regularization of SGD to identify noisy examples. We note that \textsc{ODD} is different from hard example mining~\cite{shrivastava2016training} which prunes ``easier'' examples with lower loss; this does not remove mislabeled examples effectively. The method proposed in~\cite{northcutt2017learning} is most similar to ours in principle, but is restricted to binary classification settings.
Other approaches propose to balance the examples via a pretrained network~\cite{jiang2017mentornet}, meta learning~\cite{ren2018learning}, or surrogate loss functions~\cite{ghosh2017robust,zhang2018generalized,tanaka2018joint}. Some methods require a set of trusted examples~\cite{xiao2015learning,hendrycks2018using}. 

\textsc{ODD} has several appealing properties compared to existing methods. First, the thresholds for classifying mislabeled examples from \textsc{ODD} do not rely on estimations of the noise confusion matrix. Next, \textsc{ODD} does not require additional trusted examples. Finally, \textsc{ODD} removes potentially noisy examples on-the-fly; it has little computational overhead compared to standard SGD training.
\section{Discussion}
We have proposed \textsc{ODD}, a straightforward method for robust training with mislabeled examples. \textsc{ODD} utilizes the implicit regularization effect of stochastic gradient descent, which allows us to reason counterfactually about the loss distribution of examples with uniform label noise. Based on quantiles of this (counterfactual) distribution, we can then prune examples that would potentially harm generalization.
Empirical results demonstrate that \textsc{ODD} is able to significantly outperform related methods on a wide range of datasets with artificial and real-world mislabeled examples, maintain competitiveness with \textsc{ERM} on clean datasets, as well as detecting mislabeled examples automatically in CIFAR-100. 

The implicit regularization of stochastic gradient descent opens up other research directions for implementing robust algorithms. For example, we could consider 
removing examples not only once but multiple times, retraining from scratch with the denoised dataset, or other data-augmentation approaches such as \textit{mixup}~\cite{zhang2017mixup}. Moreover, it would be interesting to understand the \textsc{ODD} from additional theoretical viewpoints, such as the effects of large learning rates.

\clearpage
%
%
\bibliographystyle{splncs04}
\bibliography{reference}

\begin{thebibliography}{10}
\providecommand{\url}[1]{\texttt{#1}}
\providecommand{\urlprefix}{URL }
\providecommand{\doi}[1]{https://doi.org/#1}

\bibitem{arpit2017a}
Arpit, D., Jastrzkebski, S., Ballas, N., Krueger, D., Bengio, E., Kanwal, M.S.,
  Maharaj, T., Fischer, A., Courville, A., Bengio, Y., Lacoste-Julien, S.: A
  closer look at memorization in deep networks. arXiv preprint arXiv:1706.05394
   (June 2017)

\bibitem{brodley1996identifying}
Brodley, C.E., Friedl, M.A., {Others}: Identifying and eliminating mislabeled
  training instances. In: Proceedings of the National Conference on Artificial
  Intelligence. pp. 799--805 (1996)

\bibitem{chaudhari2016entropy}
Chaudhari, P., Choromanska, A., Soatto, S., LeCun, Y., Baldassi, C., Borgs, C.,
  Chayes, J., Sagun, L., Zecchina, R.: {Entropy-SGD}: Biasing gradient descent
  into wide valleys. arXiv preprint arXiv:1611.01838  (November 2016)

\bibitem{chaudhari2018stochastic}
Chaudhari, P., Soatto, S.: Stochastic gradient descent performs variational
  inference, converges to limit cycles for deep networks. In: 2018 Information
  Theory and Applications Workshop (ITA). pp. 1--10. IEEE (2018)

\bibitem{cretu2008casting}
Cretu, G.F., Stavrou, A., Locasto, M.E., Stolfo, S.J., Keromytis, A.D.: Casting
  out demons: Sanitizing training data for anomaly sensors. In: Security and
  Privacy, 2008. SP 2008. IEEE Symposium on. pp. 81--95. IEEE (2008)

\bibitem{du2018algorithmic}
Du, S.S., Hu, W., Lee, J.D.: Algorithmic regularization in learning deep
  homogeneous models: Layers are automatically balanced. arXiv preprint
  arXiv:1806.00900  (June 2018)

\bibitem{ghosh2017robust}
Ghosh, A., Kumar, H., Sastry, P.S.: Robust loss functions under label noise for
  deep neural networks. In: {AAAI}. pp. 1919--1925 (2017)

\bibitem{goyal2017accurate}
Goyal, P., Doll{\'a}r, P., Girshick, R., Noordhuis, P., Wesolowski, L., Kyrola,
  A., Tulloch, A., Jia, Y., He, K.: Accurate, large minibatch {SGD}: Training
  {ImageNet} in 1 hour. arXiv preprint arXiv:1706.02677  (June 2017)

\bibitem{guo2018curriculumnet}
Guo, S., Huang, W., Zhang, H., Zhuang, C., Dong, D., Scott, M.R., Huang, D.:
  Curriculumnet: Weakly supervised learning from large-scale web images. In:
  Proceedings of the European Conference on Computer Vision (ECCV). pp.
  135--150 (2018)

\bibitem{he2015deep}
He, K., Zhang, X., Ren, S., Sun, J.: Deep residual learning for image
  recognition. arXiv preprint arXiv:1512.03385  (December 2015)

\bibitem{he2015delving}
He, K., Zhang, X., Ren, S., Sun, J.: Delving deep into rectifiers: Surpassing
  {Human-Level} performance on {ImageNet} classification. arXiv preprint
  arXiv:1502.01852  (February 2015)

\bibitem{hendrycks2018using}
Hendrycks, D., Mazeika, M., Wilson, D., Gimpel, K.: Using trusted data to train
  deep networks on labels corrupted by severe noise. arXiv preprint
  arXiv:1802.05300  (February 2018)

\bibitem{hochreiter1995simplifying}
Hochreiter, S., Schmidhuber, J.: {SIMPLIFYING} {NEURAL} {NETS} {BY}
  {DISCOVERING} {FLAT} {MINIMA}. In: Tesauro, G., Touretzky, D.S., Leen, T.K.
  (eds.) Advances in Neural Information Processing Systems 7, pp. 529--536. MIT
  Press (1995)

\bibitem{hoffer2017train}
Hoffer, E., Hubara, I., Soudry, D.: Train longer, generalize better: closing
  the generalization gap in large batch training of neural networks. In: Guyon,
  I., Luxburg, U.V., Bengio, S., Wallach, H., Fergus, R., Vishwanathan, S.,
  Garnett, R. (eds.) Advances in Neural Information Processing Systems 30, pp.
  1731--1741. Curran Associates, Inc. (2017)

\bibitem{ioffe2015batch}
Ioffe, S., Szegedy, C.: Batch normalization: Accelerating deep network training
  by reducing internal covariate shift. arXiv preprint arXiv:1502.03167  (2015)

\bibitem{jiang2017mentornet}
Jiang, L., Zhou, Z., Leung, T., Li, L.J., Fei-Fei, L.: {MentorNet}: Learning
  {Data-Driven} curriculum for very deep neural networks on corrupted labels.
  arXiv preprint arXiv:1712.05055  (December 2017)

\bibitem{keskar2016on}
Keskar, N.S., Mudigere, D., Nocedal, J., Smelyanskiy, M., Tang, P.T.P.: On
  {Large-Batch} training for deep learning: Generalization gap and sharp
  minima. arXiv preprint arXiv:1609.04836  (September 2016)

\bibitem{kleinberg2018an}
Kleinberg, R., Li, Y., Yuan, Y.: An alternative view: When does {SGD} escape
  local minima? arXiv preprint arXiv:1802.06175  (February 2018)

\bibitem{krishna2016embracing}
Krishna, R.A., Hata, K., Chen, S., Kravitz, J., Shamma, D.A., Fei-Fei, L.,
  Bernstein, M.S.: Embracing error to enable rapid crowdsourcing. In:
  Proceedings of the 2016 {CHI} Conference on Human Factors in Computing
  Systems. pp. 3167--3179. CHI '16, ACM, New York, NY, USA (2016).
  \doi{10.1145/2858036.2858115}

\bibitem{lee2018cleannet}
Lee, K.H., He, X., Zhang, L., Yang, L.: Cleannet: Transfer learning for
  scalable image classifier training with label noise. In: Proceedings of the
  IEEE Conference on Computer Vision and Pattern Recognition. pp. 5447--5456
  (2018)

\bibitem{li2017webvision}
Li, W., Wang, L., Li, W., Agustsson, E., Van~Gool, L.: {WebVision} database:
  Visual learning and understanding from web data. arXiv preprint
  arXiv:1708.02862  (August 2017)

\bibitem{li2017algorithmic}
Li, Y., Ma, T., Zhang, H.: Algorithmic regularization in over-parameterized
  matrix sensing and neural networks with quadratic activations. arXiv preprint
  arXiv:1712.09203  (December 2017)

\bibitem{liu2014classification}
Liu, T., Tao, D.: Classification with noisy labels by importance reweighting.
  arXiv preprint arXiv:1411.7718  (November 2014)

\bibitem{loshchilov2016sgdr}
Loshchilov, I., Hutter, F.: {SGDR}: Stochastic gradient descent with warm
  restarts. arXiv preprint arXiv:1608.03983  (August 2016)

\bibitem{luo2019simple}
Luo, Y., Zhu, J., Pfister, T.: A simple yet effective baseline for robust deep
  learning with noisy labels. arXiv preprint arXiv:1909.09338  (2019)

\bibitem{mahajan2018exploring}
Mahajan, D., Girshick, R., Ramanathan, V., He, K., Paluri, M., Li, Y.,
  Bharambe, A., van~der Maaten, L.: Exploring the limits of weakly supervised
  pretraining. arXiv preprint arXiv:1805.00932  (May 2018)

\bibitem{mandt2017stochastic}
Mandt, S., Hoffman, M.D., Blei, D.M.: Stochastic gradient descent as
  approximate bayesian inference. The Journal of Machine Learning Research
  \textbf{18}(1),  4873--4907 (2017)

\bibitem{neyshabur2017implicit}
Neyshabur, B.: Implicit regularization in deep learning. arXiv preprint
  arXiv:1709.01953  (September 2017)

\bibitem{northcutt2017learning}
Northcutt, C.G., Wu, T., Chuang, I.L.: Learning with confident examples: Rank
  pruning for robust classification with noisy labels. arXiv preprint
  arXiv:1705.01936  (May 2017)

\bibitem{patrini2017making}
Patrini, G., Rozza, A., Menon, A.K., Nock, R., Qu, L.: Making deep neural
  networks robust to label noise: A loss correction approach. In: 2017 IEEE
  Conference on Computer Vision and Pattern Recognition (CVPR). pp. 2233--2241.
  IEEE (2017)

\bibitem{reed2014training}
Reed, S., Lee, H., Anguelov, D., Szegedy, C., Erhan, D., Rabinovich, A.:
  Training deep neural networks on noisy labels with bootstrapping. arXiv
  preprint arXiv:1412.6596  (December 2014)

\bibitem{ren2018learning}
Ren, M., Zeng, W., Yang, B., Urtasun, R.: Learning to reweight examples for
  robust deep learning. arXiv preprint arXiv:1803.09050  (March 2018)

\bibitem{rolnick2017deep}
Rolnick, D., Veit, A., Belongie, S., Shavit, N.: Deep learning is robust to
  massive label noise. arXiv preprint arXiv:1705.10694  (May 2017)

\bibitem{russakovsky2015imagenet}
Russakovsky, O., Deng, J., Su, H., Krause, J., Satheesh, S., Ma, S., Huang, Z.,
  Karpathy, A., Khosla, A., Bernstein, M., Berg, A.C., Fei-Fei, L.: {ImageNet}
  large scale visual recognition challenge. International journal of computer
  vision  \textbf{115}(3),  211--252 (December 2015).
  \doi{10.1007/s11263-015-0816-y}

\bibitem{shrivastava2016training}
Shrivastava, A., Gupta, A., Girshick, R.: Training region-based object
  detectors with online hard example mining. In: Proceedings of the IEEE
  Conference on Computer Vision and Pattern Recognition. pp. 761--769 (2016)

\bibitem{sukhbaatar2014training}
Sukhbaatar, S., Bruna, J., Paluri, M., Bourdev, L., Fergus, R.: Training
  convolutional networks with noisy labels. arXiv preprint arXiv:1406.2080
  (June 2014)

\bibitem{sun2017revisiting}
Sun, C., Shrivastava, A., Singh, S., Gupta, A.: Revisiting unreasonable
  effectiveness of data in deep learning era. arXiv preprint arXiv:1707.02968
  (July 2017)

\bibitem{szegedy2016inception}
Szegedy, C., Ioffe, S., Vanhoucke, V., Alemi, A.: Inception-v4,
  {Inception-ResNet} and the impact of residual connections on learning. arXiv
  preprint arXiv:1602.07261  (February 2016)

\bibitem{tanaka2018joint}
Tanaka, D., Ikami, D., Yamasaki, T., Aizawa, K.: Joint optimization framework
  for learning with noisy labels. arXiv preprint arXiv:1803.11364  (2018)

\bibitem{veit2017learning}
Veit, A., Alldrin, N., Chechik, G., Krasin, I., Gupta, A., Belongie, S.J.:
  Learning from noisy {Large-Scale} datasets with minimal supervision. In:
  {CVPR}. pp. 6575--6583 (2017)

\bibitem{xiao2015learning}
Xiao, T., Xia, T., Yang, Y., Huang, C., Wang, X.: Learning from massive noisy
  labeled data for image classification. In: Proceedings of the IEEE Conference
  on Computer Vision and Pattern Recognition. pp. 2691--2699 (2015)

\bibitem{xu2019l_dmi}
Xu, Y., Cao, P., Kong, Y., Wang, Y.: L\_dmi: An information-theoretic
  noise-robust loss function. arXiv preprint arXiv:1909.03388  (2019)

\bibitem{yao2019safeguarded}
Yao, J., Wu, H., Zhang, Y., Tsang, I.W., Sun, J.: Safeguarded dynamic label
  regression for noisy supervision. AAAI (2019)

\bibitem{zagoruyko2016wide}
Zagoruyko, S., Komodakis, N.: Wide residual networks. arXiv preprint
  arXiv:1605.07146  (May 2016)

\bibitem{zhang2016understanding}
Zhang, C., Bengio, S., Hardt, M., Recht, B., Vinyals, O.: Understanding deep
  learning requires rethinking generalization. arXiv preprint arXiv:1611.03530
  (November 2016)

\bibitem{zhang2017mixup}
Zhang, H., Cisse, M., Dauphin, Y.N., Lopez-Paz, D.: mixup: Beyond empirical
  risk minimization (October 2017)

\bibitem{zhang2018generalized}
Zhang, Z., Sabuncu, M.: Generalized cross entropy loss for training deep neural
  networks with noisy labels. In: Advances in neural information processing
  systems. pp. 8778--8788 (2018)

\end{thebibliography}

\appendix
\newpage
\onecolumn
\appendix

\section{Additional Experimental Results}
\subsection{Ablation Studies}
\label{sec:exp-ablation}


\subsubsection{Sensitivity to $p$} 
\begin{figure}[h]
    \centering
    \includegraphics[width=0.4\textwidth]{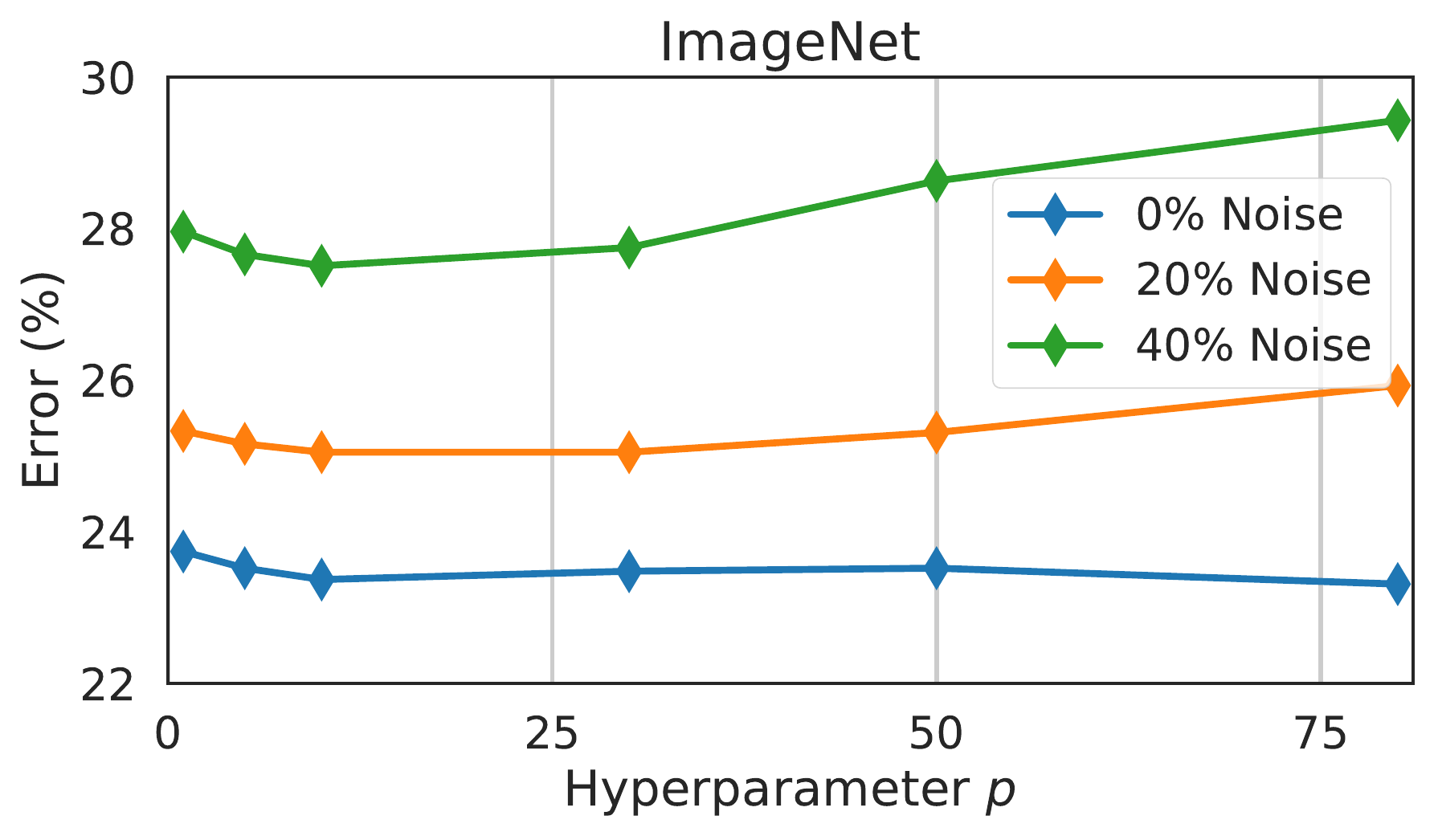}
    \caption{Ablation studies over the hyperparameter $p$ on ImageNet under different levels of mislabeled examples.}
    \label{fig:app:ablation-imagenet-p}
\end{figure}

We first evaluate noisy ImageNet classification with varying $p$. 
A higher $p$ includes more clean examples at the cost of involving more noisy examples. 
From Figure~\ref{fig:app:ablation-imagenet-p}, \textsc{ODD} is not very sensitive to $p$, and empirically $p = 10$ represents the best trade-off. 


\subsubsection{Sensitivity to $E$} 
\begin{figure}[h]
\centering
\includegraphics[width=0.50\textwidth]{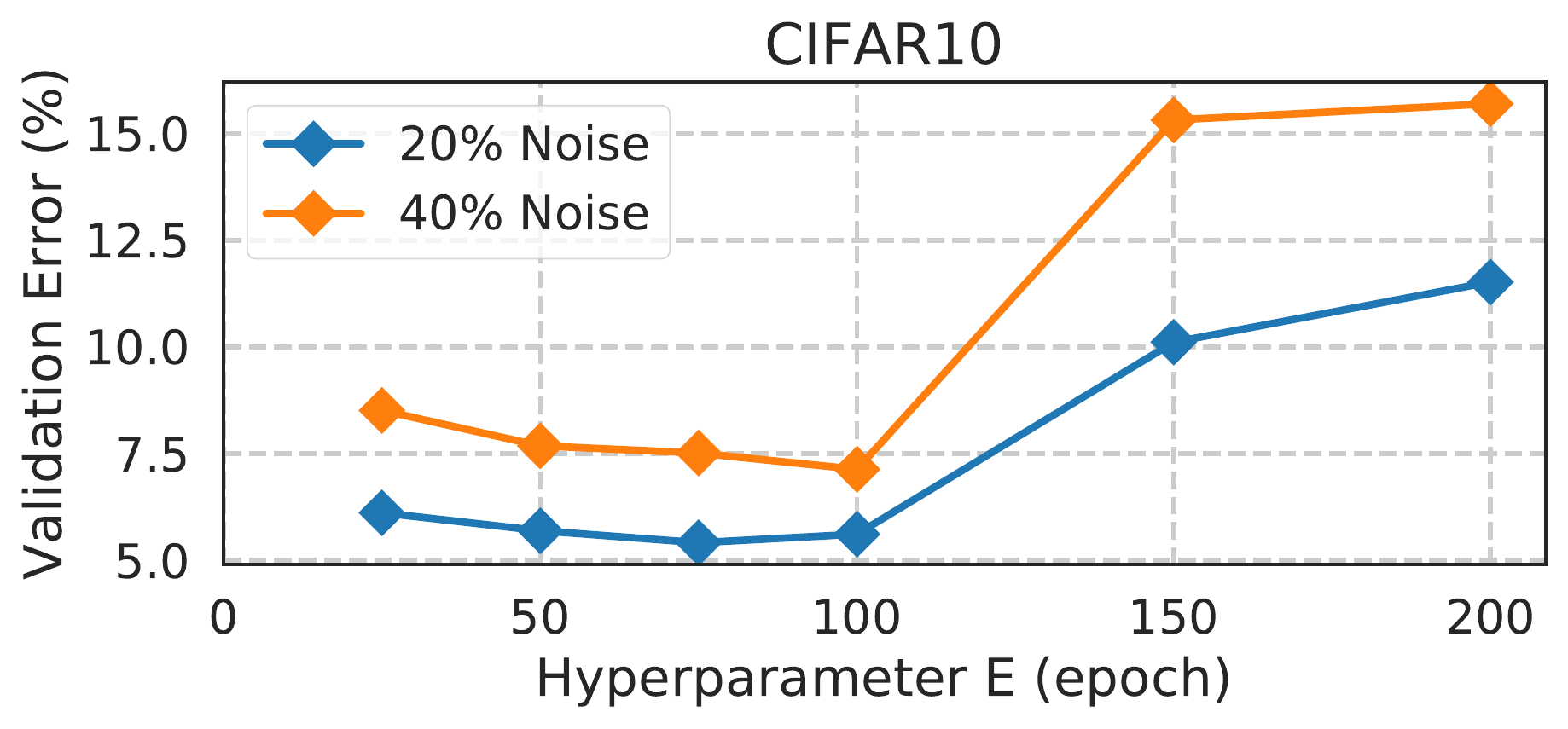}
\caption{Validation errors of \textsc{ODD} on CIFAR10 with different values of $E$.
}
\label{fig:app:epoch-vs-val-error}
\end{figure}

We evaluate the validation error of \textsc{ODD} on CIFAR with $20\%$ and $40\%$ input-agnoistic label noise where $E \in \{25, 50, 75, 100, 150, 200\}$ ($E = 200$ is equivalent to \textsc{ERM}). The results in Figure~\ref{fig:app:epoch-vs-val-error} suggest that our method is able to separate noisy and clean examples if $E$ is relatively small where the learning rate is high, but is unable to perform well when the learning rate decreases.

\subsubsection{Sensitivity to the amount of noise} 
\begin{figure}
\centering
\includegraphics[width=0.50\textwidth]{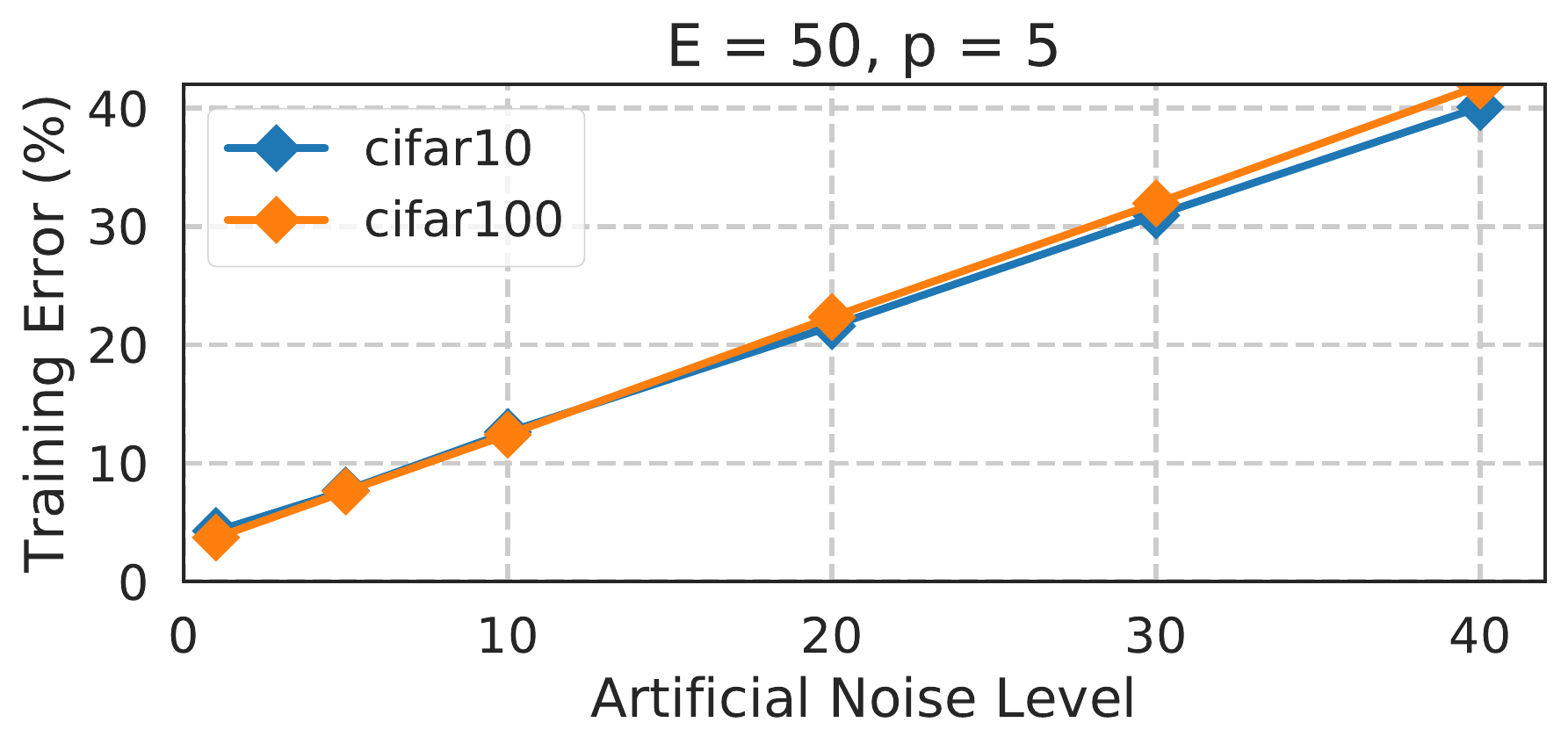}
\caption{Training errors of \textsc{ODD} on CIFAR10 with different amount of uniform noise.
}
\label{fig:app:noise-vs-train-error}
\end{figure}

Finally, we evaluate the training error of \textsc{ODD} on CIFAR under input-agnostic label noise of $\{1\%, 5\%, 10\%, 20\%, 30\%, 40\%\}$ with $p = 5$, $E = 50$ or $75$. This reflects how much examples exceed the threshold and are identified as noise at epoch $E$. From Figure~\ref{fig:app:noise-vs-train-error}, we observe that the training error is almost exactly the amount of noise in the dataset, which demonstrates that the loss distribution of noise can be characterized by our threshold regardless of the percentage of noise in the dataset.

\subsubsection{Precision and recall for classifying noise} 
\begin{figure*}[htbp]
\centering
\includegraphics[width=0.50\textwidth]{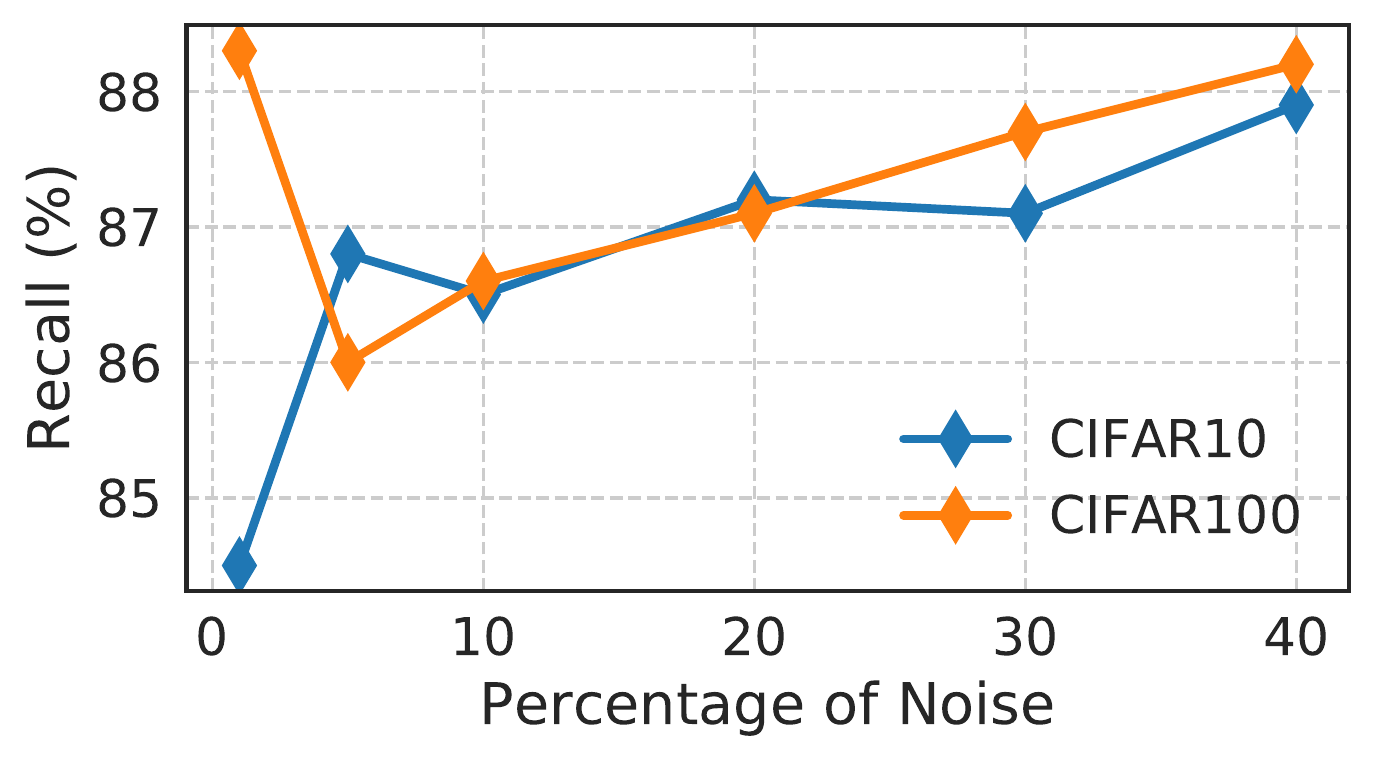}
\hspace*{0.05\textwidth}
\includegraphics[width=0.50\textwidth]{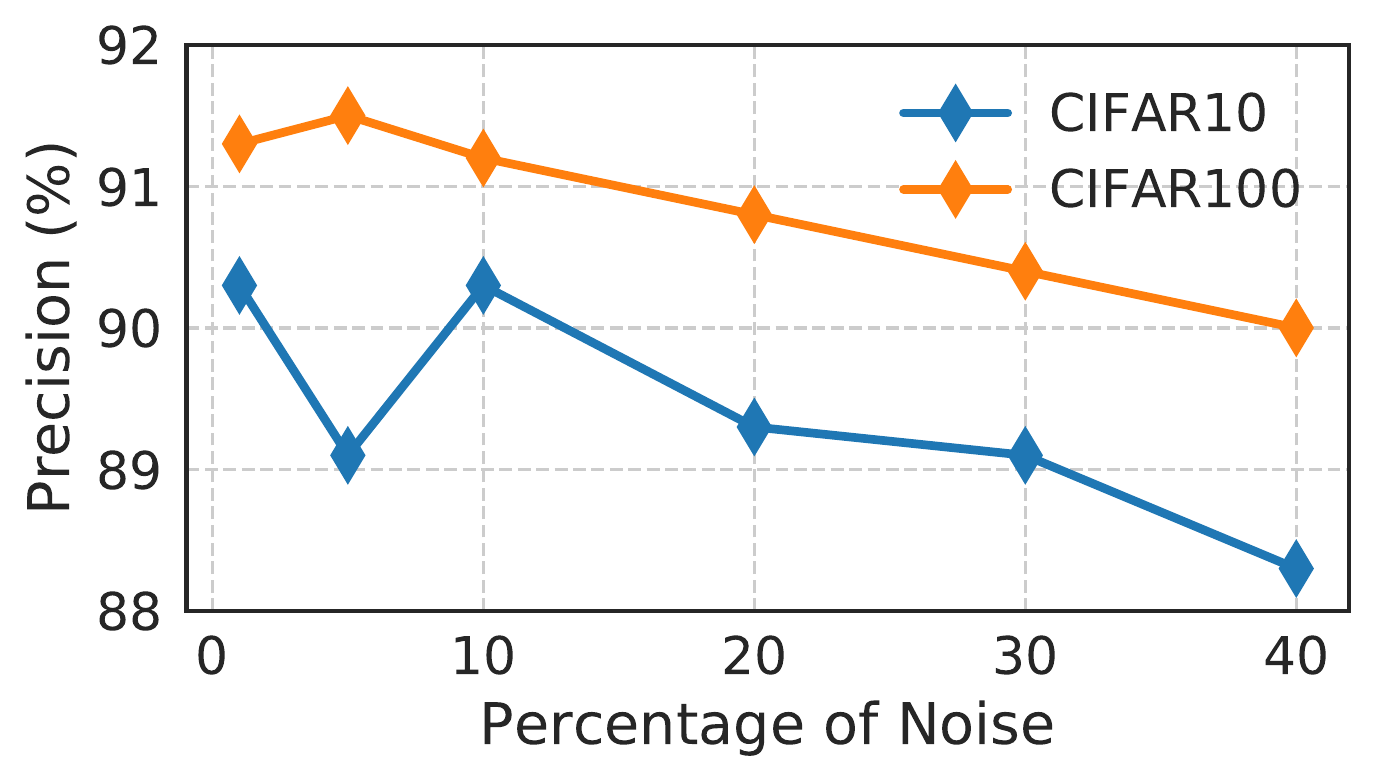}
\caption{Recall and precision for \textsc{ODD} on CIFAR10 and CIFAR100 with different levels of uniform random noise. 
}
\label{fig:app:cifar-prec-recall}
\end{figure*}

We evaluate precision and recall for examples classified as noise on CIFAR10 and CIFAR100 for different noise levels (1, 5, 10, 20, 30, 40) in Figure~\ref{fig:app:cifar-prec-recall}. The recall values are around 0.84 to 0.88 where as the precision values range from 0.88 to 0.92. This demonstrates that \textsc{ODD} is able to achieve good precision/recall with default hyperparameters even at different noise levels.

\subsubsection{Percentage of samples discared by \textsc{ODD}} We show the percentage of examples discarded by \textsc{Noise Classifier} in Table~\ref{tab:ablation-imagenet-discard}; the percentage of discarded examples by $p = 10$ is very close to the actual noise level, suggesting that it is a reasonable setting. 

\begin{table*}[htbp]
\begin{center}
\caption{Percentage of example discraded by \textsc{ODD} on ImageNet-2012.}
\label{tab:ablation-imagenet-discard}
\begin{tabular}{c|ccccc|c}
\toprule
\multirow{2}{*}{\% Mislabeled}  & \multicolumn{5}{|c|}{Hyperparameter $p$} & \multirow{2}{*}{Network}
\\
& 1 & 10 & 30 & 50 & 80 & \\\midrule
$0\%$ & 5.5 & 2.3 & 1.1 & 0.7 & 0.4 & \multirow{3}{*}{ResNet-152} \\
$20\%$ & 23.8 & 20.8 & 19.2 & 17.5 & 0.7 &\\
$40\%$ & 44.1 & 40.2 & 36.2 & 27.6 & 0.6  & \\
\bottomrule
\end{tabular}
\end{center}
\end{table*}

\subsubsection{Ablation studies on WebVision} We include additional ablation on $p$ for WebVision (Table~\ref{tab:wv}). While the results for $p = 30$ is slightly better, our method outperforms other methods (\textsc{Luo}) even with worse hyperparameters.

\begin{table}
\centering
\caption{Additional results on WebVision with varying $p$.}
\vspace{1em}
\begin{tabular}{cllll}
\toprule
 $p$ & \multicolumn{2}{l}{Webvision}      & \multicolumn{2}{l}{ImageNet}                 \\ \midrule
                   & Top1      & Top5  & Top1  & Top 5\\\midrule
1               & 74.01     & 89.93 & 65.77    & 85.40        \\
10              & 74.31     & 90.55 & 66.09    & 85.86       \\
30              & 74.62     & 90.63 & 66.73    & 86.32        \\
50              & 74.43     & 90.78 & 66.58    & 86.21        \\
80              & 74.33      & 90.30 & 66.23    & 86.24       \\ \bottomrule
\end{tabular}
\label{tab:wv}
\end{table}

\subsection{Images in CIFAR-100 Classified as Noise}
\label{sec:app:noisy-cifar}
\begin{figure*}[htbp]
    \centering
    \includegraphics[width=0.9\textwidth]{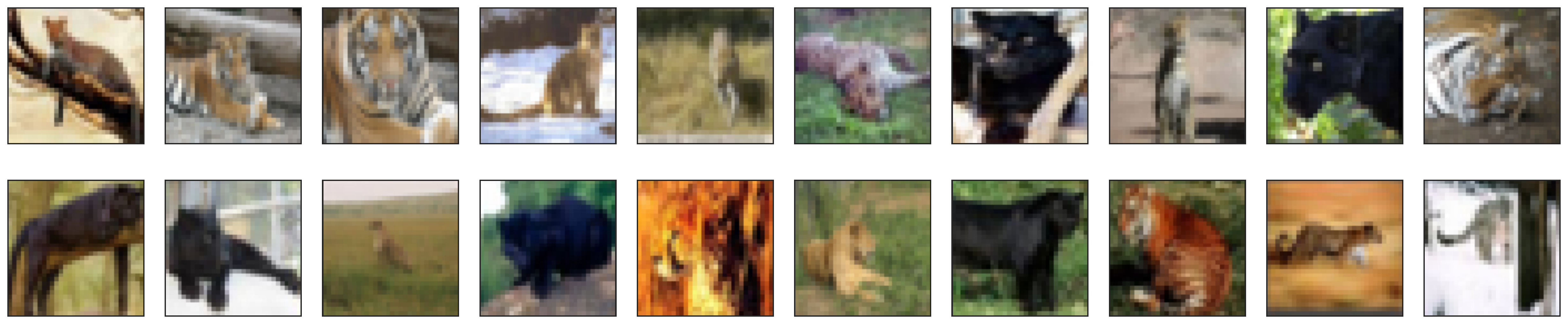}
    \caption{Examples with label ``leopard'' that are classified as noise.}
    \label{fig:app:leopard}
\end{figure*}

We display the examples in CIFAR-100 training set for which our \textsc{ODD} methods identify as noise across 3 random seeds. One of the most common label such examples have is ``leopard''; in fact, 21 of 50 ``leopard'' examples in the training set are perceived as hard, and we show some of them in Figure~\ref{fig:app:leopard}. It turns out that a lot of the ``leopard'' examples contains images that clearly contains tigers and black panthers (CIFAR-100 has a label corresponding to ``tiger'').

\begin{figure*}[htbp]
    \centering
    \includegraphics[width=0.9\textwidth]{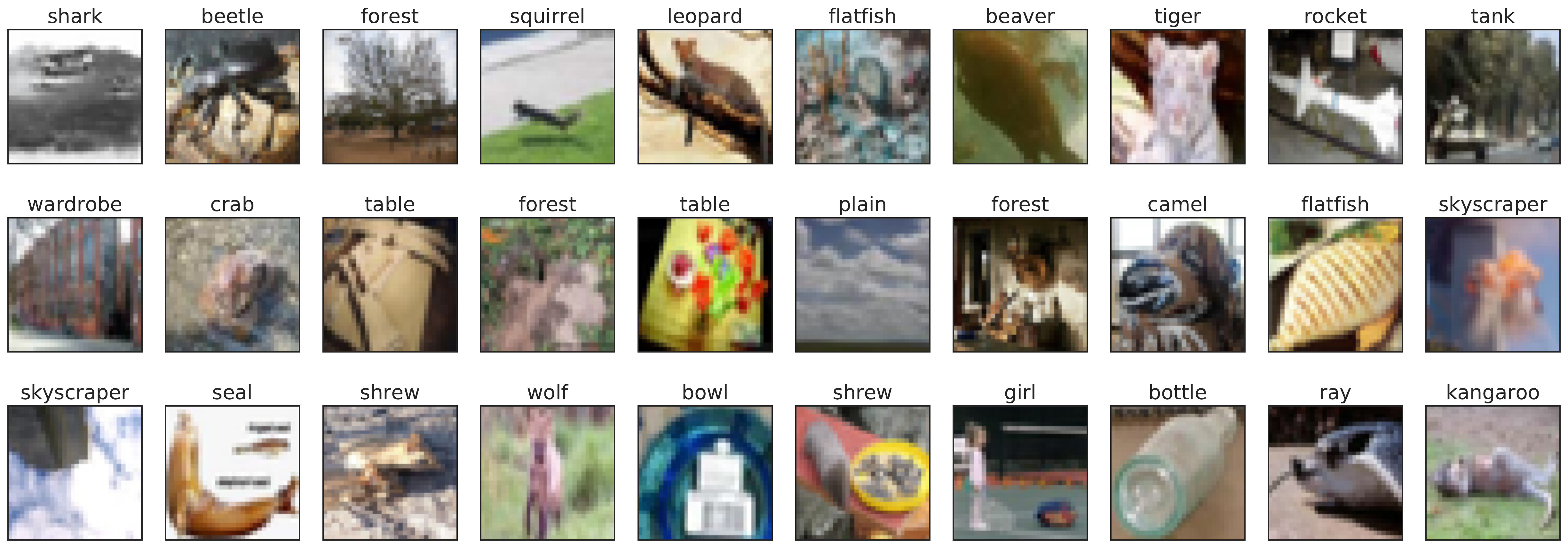}
    \caption{Random CIFAR-100 examples that are classified as noise.}
    \label{fig:app:noisy-images-cifar100}
\end{figure*}

\begin{figure*}[htbp]
    \centering
    \includegraphics[width=0.9\textwidth]{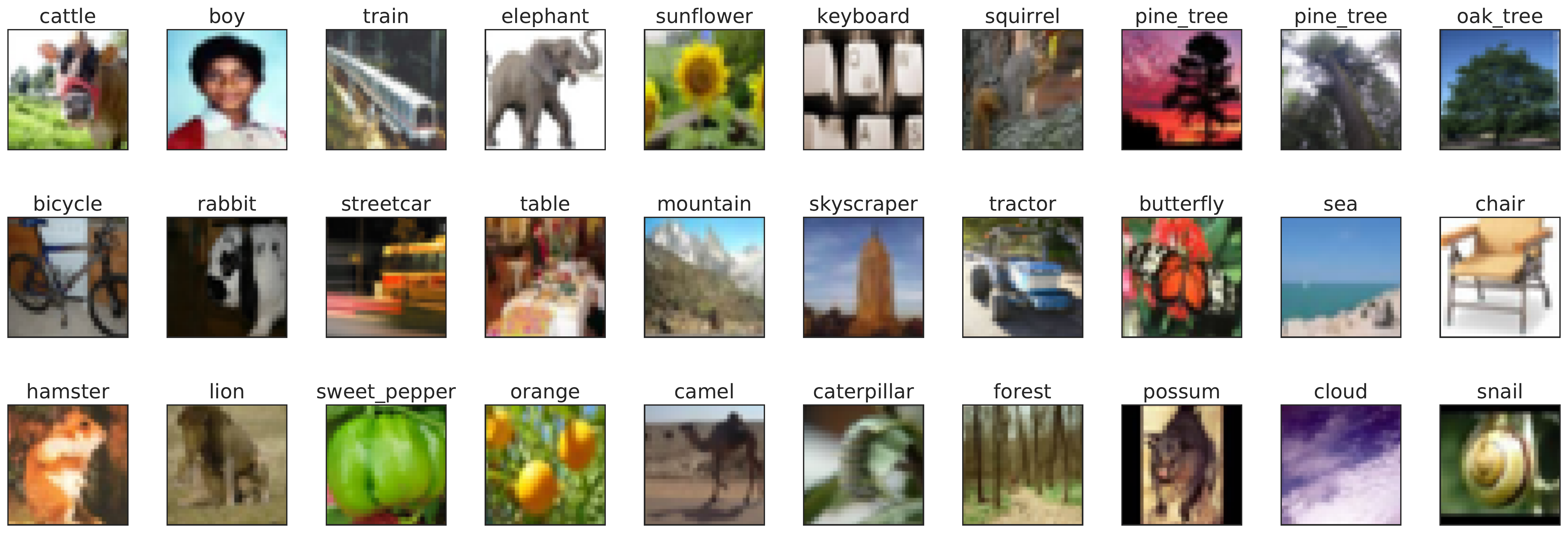}
    \caption{Random CIFAR-100 examples that are not classified as noise.}
    \label{fig:app:clean-images-cifar100}
\end{figure*}

We also demonstrate random examples from the CIFAR-100 that are identified as noise in Figure~\ref{fig:app:noisy-images-cifar100} and those that are not identified as noise in Figure~\ref{fig:app:clean-images-cifar100}. The examples identified as noise often contains multiple objects, and those not identified as noise often contains only one object that is less ambiguous in terms of identity.

\end{document}